\documentclass[11pt,a4paper]{article}

\usepackage[hyperref]{acl2021}
\usepackage{times}
\usepackage{latexsym}

\usepackage{microtype}

\aclfinalcopy 

\usepackage[T1]{fontenc}

\usepackage{mathtools}
\usepackage{amsthm}
\usepackage{amsmath}
\usepackage{amssymb}
\usepackage{mathrsfs}
\usepackage{newtxtext, newtxmath}
\usepackage{tikz}
\usepackage{tikz-cd}
\usepackage{tikzit}
\usepackage{subcaption}

\newcommand{\s}{\enspace}
\newcommand{\sub}{\subseteq}
\newcommand{\size}[1]{\left\vert{#1}\right\vert}

\newcommand{\bb}[1]{\mathbb{#1}}
\renewcommand{\phi}{\varphi}

\newcommand{\N}{\bb{N}}
\newcommand{\Z}{\bb{Z}}
\newcommand{\R}{\bb{R}}

\newcommand{\G}{\mathbf{G}}
\newcommand{\Vect}{\mathbf{Vect}}

\DeclareMathOperator*{\argmin}{arg\,min}

\def\num#1{\numx#1}\def\numx#1e#2{{#1}\mathrm{e}{#2}}

\def\then{\mathbin{\raise 0.6ex\hbox{\oalign{\hfil$\scriptscriptstyle      \mathrm{o}$\hfil\cr\hfil$\scriptscriptstyle\mathrm{9}$\hfil}}}}

\newtheorem{definition}{Definition}[section]
\newtheorem{remark}[definition]{Remark}
\newtheorem{example}[definition]{Example}

\title{Functorial Language Models\\ {\large (Work In Progress)}}

\author{Alexis Toumi \\
  University of Oxford \\
  Cambridge Quantum Computing Ltd. \\
  \texttt{alexis@toumi.email} \\ \And
  Alex Koziell-Pipe \\
  \texttt{alex.koziellpipe@gmail.com} \\ }

\begin{document}
\maketitle


\section*{Introduction}

We introduce functorial language models: a principled way to compute
probability distributions over word sequences given a monoidal functor from
grammar to meaning. This yields a method for training categorical
compositional distributional (DisCoCat) models on raw text data.
We provide a proof-of-concept implementation in DisCoPy \cite{DeFeliceEtAl20a},
see appendix~\ref{a-experiment}.

Language models, i.e. probability distributions over word sequences, are a
cornerstone of natural language processing and information retrieval \cite{PonteCroft98}.
Neural models \cite{BengioEtAl03}, where these distributions are learnt by a
neural network, are now built in everyday tools such as virtual assistants,
automatic translation, etc. However effective they may be, neural models are not
compositional in the sense that they do not take the grammatical structure of
sentences into account, at least not explicitly. This severely limits our ability
to interpret them: if we open the black box, we only see matrices of weights.

On the other hand, categorical compositional distributional (DisCoCat) models
\cite{ClarkEtAl08,ClarkEtAl10} use grammatical structure to compose the
distributional meaning of words together into a meaning for the sentence.
Grammar is explicitly represented as string diagrams, which allow formal
reasoning about natural language semantics, for example analysing ambiguity
\cite{KartsaklisEtAl13,KartsaklisEtAl14,PiedeleuEtAl15}
and entailment \cite{SadrzadehEtAl18,DeFeliceEtAl19a}.

However, despite
emprical validation on small-scale examples \cite{GrefenstetteSadrzadeh11},
applying DisCoCat models at a large scale to real-world text data appears
like a far-away goal.
The challenge is at least two-fold: 1) we need a robust way to predict the
grammatical structure of a given word sequence, 2) we need an efficient way
to learn the meaning of each word, given the grammatical structures in which
it occurs. The first point requires a probabilistic grammar: it is not enough
to know whether a sequence is grammatical, we need a probability distribution
over its possible parsings.

It has also been argued that a robust model ought to
be incremental \cite{SadrzadehEtAl18a}: scanning through the given word sequence
left-to-right and updating its prediction for what comes next. This plays a
crucial role in the context of informal language and dialogues \cite{PurverEtAl06}.
Previous work by the first author and collaborators \cite{ShieblerEtAl20}
characterise these incremental, probabilistic parsers in terms of a functor
from formal grammars to automata.

The second challenge, i.e. how to use grammar for learning the meaning of words,
is essential if we want DisCoCat to work as a standalone model for natural
language. Indeed, the first experiments \cite{GrefenstetteSadrzadeh11}
assumed the vector embedding for nouns to be given by some other means, such
as co-occurence counts or neural language models. On the other hand, in more
recent experiments on quantum hardware \cite{MeichanetzidisEtAl20a,
MeichanetzidisEtAl20,CoeckeEtAl20} the meaning of words is learnt directly from
the training data of a supervised question-answering task.
We coin the term \emph{functorial learning} for this approach, as it can be
understood as learning a functor from data.

We give a formal definition of this functorial approach and define \emph{
functorial language models} by composition with a probabilistic grammar.
Concretely, we train a DisCoCat model to predict which word is missing in a
sentence with a hole. We argue that this captures precisely the idea of
distributional compositionality: extending Firth's principle, we shall know a
word by the company it keeps and by the grammatical structure in which it occurs.


\section{Functorial learning for DisCoCat}

DisCoCat models have a one-sentence definition: they are rigid monoidal functors
from the category generated by a pregroup grammar to vector spaces\footnote{
We only consider finite dimensional vector spaces.} and linear maps.
Let's unpack this definition and repack it in a format suitable for computation.

Fix a set of words $V$, called the vocabulary, and a set of basic types $X$.
A rule is a pair of sequences of words and types
$r \in (V + X)^\star \times (V + X)^\star$, where $\times$, $+$ and ${}^\star$
denote respectively the Cartesian product, the disjoint union and the free
monoid $X^\star = \coprod_{n \in \N} X^n$ with unit $1$ the empty string.
A grammar $G = (V, X, R, s)$ is given by a set $R$ of such rules together with
a distinguished $s \in X$ called the sentence type.
This defines a signature in the sense of \cite{Selinger10} which generates
a free monoidal category $\G$. The objects are sequences of words and
types, the arrows are progressive\footnote{A diagram is progressive when its
wires go monotonously top to bottom, i.e. they do not bend up or down.}
planar\footnote{A diagram is planar when its wires do not cross.} string diagrams
with rules as boxes. The language of $G$ is defined as the sequences
$x \in V^\star$ such that there is a diagram $g : x \to s$ in $\G$:
its grammatical structure.

This definition merely reformulates the notion of semi-Thue system \cite{Thue14}
in the language of monoidal category theory. \cite{Post47} and \cite{Markov47}
independently proved the undecidability of the parsing problem, i.e. given
$G$ and $x \in V^\star$ decide whether there exists $g  : x \to s$ in $\G$.
\cite{Chomsky57} then put the equivalent notion of unrestricted grammar at
the bottom of his well-known hierarchy. In parallel, \cite{Lambek58} defined
his calculus in terms of closed monoidal categories, taking inspiration from the
categorial grammars of \cite{Adjukiewicz35} and \cite{Bar-Hillel53}. Half a
century later, pregroup grammars \cite{Lambek99,Lambek01,Lambek08} simplified
this calculus by going from closed to rigid monoidal categories. Pregroup
grammars are at the basis of the original DisCoCat model of \cite{ClarkEtAl08}.

\begin{definition}
A pregroup grammar has types $X \times \Z$ where for $n \in \N$ the types
$(x, -n)$ and $(x, +n)$ are written $x^{l...l}$ and $x^{r...r}$ respectively
and are called iterated left and right adjoints.
The rules are of two kinds: cups and triangles. Cups have the shape $(x, z)
(x, z + 1) \to 1$ for $z \in \Z$, i.e. they cancel a basic type with its right
adjoint. Triangles have the shape $w \to t$ for a word $w \in V$ and a type
$t \in (X \times \Z)^\star$. Note that we draw words as the
labels of triangle boxes rather than as input wires.
\end{definition}

A pregroup grammar $G = (V, X, R, s)$ defines a rigid signature: the generating
objects are given by words and types $V + X$, the generating arrows are given
by dictionary entries $w \to t \in R$. Thus, it generates a free rigid monoidal
category $\G$ where the object are sequences of words and types with iterated
adjoints, the arrows are planar string diagrams with triangles as boxes.
Cups are given by the rigid structure of $\G$, they are drawn as bent wires.
Again, the language of $G$ is defined as
$\{ x \in V^\star \ \vert \ \exists \ g : x \to s \in \G \}$.

The main distinction between Chomsky's phrase structure grammars
and the categorial grammar tradition of Adjukiewicz, Bar-Hillel and Lambek
is that in the latter ``all the grammar is in the lexicon''. The only
language-dependent rules are dictionary entries of the form $r : w \to t$ for a
word $w \in V$ and a type $t \in \text{Ob}(\G)$. They are drawn as triangle boxes.
All the grammatical rules come from the structure of the category $\G$,
e.g. the cups of pregroups from the rigid structure.
This has both conceptual and computational advantages. Conceptually, it makes
categorial grammars universal: the same rules apply to all languages,
only the dictionaries change. Computationally, these universal rules have a
canonical semantics, thus it is enough to define the meaning of each dictionary
entry.

\begin{definition}
A DisCoCat model is a rigid monoidal functor $F : \G \to \Vect$ from the
category generated by a pregroup grammar. On objects $F$ is
defined by a mapping $F_0 : X \to \N$ which sends each basic type to the
dimension of a vector space. The image of word types $w \in V$ is the monoidal
unit $F(w) \!=\! 1$, the image of complex types $t \!=\! (x_1, z_1) \!\dots\! (x_n, z_n)$ is
the product of the dimensions of their basic types
$F(t) \!=\! F_0(x_1) \!\dots\! F_0(x_n)$.
On arrows $F$ is defined by a mapping $F_1 \!\in\! \coprod_{(w, t) \in R} \R^{F(t)}$
that sends each dictionary entry $(w, t) \in R$ to a vector $F(w, t) \in \R^n$
for $n = F(t)$ the dimension of its type. Cups
are sent to the rigid monoidal structure of $\Vect$. The meaning of a sentence
$g : x \to s$ is given by the vector $F(g) \in \R^m$ for $m = F(s)$ the
dimension of the sentence type. When $F(s) = 1$, the meaning of sentences is
given by real-valued scalars that can encode either truth value or likelihood.
\end{definition}

\begin{remark}
DisCoCat models have been generalised both in their domain and codomain: one can
vary the grammar category as in \cite{CoeckeEtAl13} or the semantic category
as in \cite{PiedeleuEtAl15,CoeckeEtAl18,Delpeuch19}.
While we stick to the original definition, our proposal can apply
to any model, so long as the meaning of words lives in some vector space,
regardless of how these word vectors are composed together.
\end{remark}

We fix the map $F_0 : X \to \N$ and consider it as the hyper-parameters of the
model. All the data defining the DisCoCat model is contained in the finite
set of vectors $F_1 \in \coprod_{(w, t) \in R} \R^{F(t)}$. Thus, we may consider
these $D = \sum_{(w, t) \in R} F(t)$ parameters as a machine learning landscape.
The idea of functorial learning is to take a training set of pairs
$X \sub Ar(\G) \times Ar(\Vect)$ of diagrams $d \in Ar(\G)$ and vectors
$y \in Ar(\Vect)$, and learn a functor $F$ such that $F(d) = y$.
In practice, this exact functor may not exist thus we fix a loss
$L : \coprod_{n \in \N} \R^n \times \R^n \to \R^+$ and a regularisation
$R :\R^D \to \R^+$ then using gradient-based methods we approximate:
$$\argmin_{F : \G \to \Vect} \quad
R(F_1) \s + \sum_{(d, y) \in X} L\big(F(d), y\big)$$

\begin{example}
Fix a type $q \in X$ for yes-no questions with $F(q) = 1$, i.e. the meaning of a
question $d : w_1 ... w_n \to q$ is a scalar $F(d) \in \R$. Take a training set
of such questions together with their answer in $\{ 0, 1 \} \sub \R$. Then the
functor $F$ may be seen as a DisCoCat model for question-answering, see \cite{
DeFeliceEtAl19a}. This model has been deployed on quantum hardware, see \cite{
MeichanetzidisEtAl20a,MeichanetzidisEtAl20,CoeckeEtAl20}.
\end{example}


\section{Functorial language models}

The definition of DisCoCat models can be recast in terms of encoding matrices,
first introduced in \cite{Toumi18,CoeckeEtAl18a}. The set of meaning vectors
$F_1 \in \coprod_{(w, t) \in R} \R^{F(t)}$ can be packed into a set of matrices
$E_t : \size{V_t} \to F_0(t)$ indexed by grammatical types
$t \in (X \times \Z)^\star$, where $V_t \sub V$ is the set of words $w \in V$
with $(w, t) \in R$. If we compose the matrix $E_t$ with the one-hot vector for
a word $w : 1 \to \size{V_t}$, we get its image under the functor
$F(w, t) = w \then E_t$. In the other direction, if we compose a meaning (co)vector $v : F_0(t) \to 1$
with $E_t$, we get a (co)vector $E_t \then v : \size{V_t} \to 1$ which gives its
inner product with the words in $V_t$.
The key insight of our functorial language models is to take
$\text{softmax}(E_t \then v)$ as the probability distribution over words in
$V_t$ given a meaning $v : F_0(t) \to 1$.

\begin{remark}
$\text{softmax}$ is not a linear map, hence technically we cannot draw it as a
box in a $\Vect$-valued diagram. We can however draw it as a bubble, see \cite{
ToumiEtAl21}. We can then consider functors which send bubbles to $\text{softmax}$.
\end{remark}

Fix a corpus of grammatical sentences $X \sub \coprod_{x \in V^\star} \G(x, s)$
and assume $F(s) = 1$, i.e. the meaning of sentences are scalars.
We generate a training set from this by taking each sentence $d : w_1 ... w_n
\to s$, removing one dictionary entry $w_i \to t_i$ by replacing it with the
identity $t_i \to t_i$. Thus, we get a diagram $d_i : t_i \to s$ and take
the one-hot vector for the word $w_i$ as target label.
Intuitively, given the diagram of a sentence with a hole, we want
to predict what word is missing.

This does not yet define a language model in the usual sense, indeed we are
assuming that text data comes annotated with grammatical structure. That is, we
are computing the conditional distribution $P(w \vert d)$ over words
$w \in V_t$ of type $t \in \text{Ob}(\G)$ given the grammatical structure $d : t \to s$ in which
they occur. In order to get a language model, we need to compose this
conditional with a distribution $P(d \vert w_1 ... w_n)$ over diagrams
given word sequences: a probabilistic grammar. Learning this distribution from
data is called probabilistic grammar induction, which \cite{ShieblerEtAl20}
formulate in terms of monoidal categories. In future work, we plan
to train our functorial language model end-to-end on raw text data, i.e.
learning both the probabilistic grammar and the DisCoCat model at once.


\section{Implementation}

We implemented a proof-of-concept in DisCoPy \cite{DeFeliceEtAl20a} and used jax
\cite{BradburyEtAl20} for automatic differentiation and just-in-time compilation
on GPU. We initialise the encoding matrices at random
and use singular value decomposition to obtain word vectors that are close to
orthogonal. We used cross-entropy loss, a weighted sum of l$1$ and l$2$
($\num{1e-1}$ and $\num{5e-2}$ resp.) regularisation and
Adam optimization \cite{KingmaBa17} (learning rate $=\num{5e-2}$).
We set the hyper-parameters to $F(s) = 1$ and $F(n) = 7$.
After training on a subset of hand-crafted data
(\textasciitilde 100 sentences with three distinct grammatical structures),
the model was able to infer the missing word in previously unseen sentences with
accuracy $> .75$, see appendix~\ref{a-experiment}. The notebook can be found on
\href{https://github.com/oxford-quantum-group/discopy/blob/aade82b44c06eb632f75708a2e8bffd6cfae96ec/docs/notebooks/functorial_language_model.ipynb}{DisCoPy's GitHub}.

\footnotesize{\bibliographystyle{apalike}
\bibliography{semspace-language-models}}

\appendix

\section{Experiment}\label{a-experiment}

We reproduce our hand-crafted dataset together with some sample predictions from
our model. In the following cherry-picked examples, the prediction is correct in the
first three cases and incorrect in the last one. This failure can be explained
by the fact the word ``krill'' only appears once in the training set.

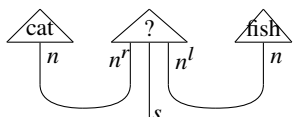
\vspace{10pt}
\begin{tikzpicture}[baseline={(0.base)}, scale=.25]
	\begin{pgfonlayer}{nodelayer}
		\node [style=none] (0) at (-3.75, 3.5) {};
		\node [style=none] (1) at (1, 4.25) {};
		\node [style=none] (2) at (1, 2) {};
		\node [style=none, right] (3) at (-0.15, 3.4) {$n^r$};
		\node [style=none] (4) at (2, 4.25) {};
		\node [style=none] (5) at (2, 0) {};
		\node [style=none, right] (6) at (2.1, 0.4) {$s$};
		\node [style=none] (7) at (3, 4.25) {};
		\node [style=none] (8) at (3, 2) {};
		\node [style=none, right] (9) at (3.35, 3.4) {$n^l$};
		\node [style=none] (10) at (-3.75, 4.25) {};
		\node [style=none] (11) at (-3.75, 2) {};
		\node [style=none, right] (12) at (-3.4, 3.4) {$n$};
		\node [style=none] (13) at (8, 4.25) {};
		\node [style=none] (14) at (8, 2) {};
		\node [style=none, right] (15) at (8.35, 3.4) {$n$};
		\node [style=none] (16) at (-1.5, 0.75) {};
		\node [style=none] (17) at (5.5, 0.75) {};
		\node [style=none] (18) at (0, 4.25) {};
		\node [style=none] (19) at (4, 4.25) {};
		\node [style=none] (20) at (2, 6) {};
		\node [style=none] (22) at (2, 5) {?};
		\node [style=none] (23) at (-5.5, 4.25) {};
		\node [style=none] (24) at (-2, 4.25) {};
		\node [style=none] (26) at (-3.75, 6) {};
		\node [style=none] (27) at (-3.75, 5) {cat};
		\node [style=none] (28) at (6.5, 4.25) {};
		\node [style=none] (29) at (9.5, 4.25) {};
		\node [style=none] (30) at (8, 6) {};
		\node [style=none] (32) at (8, 5) {fish};
	\end{pgfonlayer}
	\begin{pgfonlayer}{edgelayer}
		\draw [in=90, out=-90] (1.center) to (2.center);
		\draw [in=90, out=-90] (4.center) to (5.center);
		\draw [in=90, out=-90] (7.center) to (8.center);
		\draw [in=90, out=-90] (10.center) to (11.center);
		\draw [in=90, out=-90] (13.center) to (14.center);
		\draw [in=180, out=-90] (11.center) to (16.center);
		\draw [in=0, out=-90] (2.center) to (16.center);
		\draw [in=180, out=-90] (8.center) to (17.center);
		\draw [in=0, out=-90] (14.center) to (17.center);
		\draw [-, fill=white] (18.center) to (19.center);
		\draw [-, fill=white] (19.center) to (20.center);
		\draw [-, fill=white] (23.center) to (24.center);
		\draw [-, fill=white] (26.center) to (23.center);
		\draw [-, fill=white] (28.center) to (29.center);
		\draw [-, fill=white] (29.center) to (30.center);
		\draw (26.center) to (24.center);
		\draw (20.center) to (18.center);
		\draw (30.center) to (28.center);
	\end{pgfonlayer}
\end{tikzpicture}
\\
{\ttfamily \scriptsize
\textbf{Target:} eats\\
\textbf{Prediction:} eats (0.60), bites (0.35), flees (0.05)}
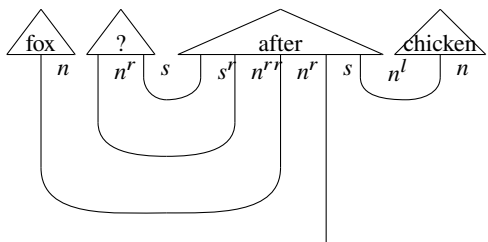
\vspace{10pt}

\begin{tikzpicture}[baseline={(0.base)}, scale=.6]
	\begin{pgfonlayer}{nodelayer}
		\node [style=none] (0) at (-0.25, 4) {};
		\node [style=none] (1) at (1, 4.25) {};
		\node [style=none] (2) at (1, 2.75) {};
		\node [style=none, right] (3) at (1.35, 3.9) {$n^r$};
		\node [style=none] (4) at (2, 4.25) {};
		\node [style=none] (5) at (2, 3.75) {};
		\node [style=none, right] (6) at (2.35, 3.9) {$s$};
		\node [style=none] (7) at (-0.25, 4.25) {};
		\node [style=none] (8) at (-0.25, 1.75) {};
		\node [style=none, right] (9) at (0.1, 3.9) {$n$};
		\node [style=none] (10) at (3.25, 4.25) {};
		\node [style=none] (11) at (3.25, 3.75) {};
		\node [style=none, right] (12) at (3.6, 3.9) {$s^r$};
		\node [style=none] (13) at (4, 4.25) {};
		\node [style=none] (14) at (4, 2.75) {};
		\node [style=none, right] (15) at (4.35, 3.9) {$n^{rr}$};
		\node [style=none] (16) at (5, 4.25) {};
		\node [style=none] (17) at (5, 1.75) {};
		\node [style=none, right] (18) at (5.35, 3.9) {$n^r$};
		\node [style=none] (19) at (6, 4.25) {};
		\node [style=none] (20) at (6, 0) {};
		\node [style=none, right] (21) at (6.35, 3.9) {$s$};
		\node [style=none] (22) at (6.75, 4.25) {};
		\node [style=none] (23) at (6.75, 3.75) {};
		\node [style=none, right] (24) at (7.35, 3.9) {$n^l$};
		\node [style=none] (25) at (8.5, 4.25) {};
		\node [style=none] (26) at (8.5, 3.75) {};
		\node [style=none, right] (27) at (8.85, 3.9) {$n$};
		\node [style=none] (28) at (2.5, 3.25) {};
		\node [style=none] (29) at (2.5, 2) {};
		\node [style=none] (30) at (2.5, 0.75) {};
		\node [style=none] (31) at (7.75, 3.25) {};
		\node [style=none] (32) at (0.75, 4.25) {};
		\node [style=none] (33) at (2.25, 4.25) {};
		\node [style=none] (34) at (1.5, 5.25) {};
		\node [style=none] (36) at (1.5, 4.5) {?};
		\node [style=none] (37) at (-1, 4.25) {};
		\node [style=none] (38) at (0.5, 4.25) {};
		\node [style=none] (39) at (-0.25, 5.25) {};
		\node [style=none] (41) at (-0.25, 4.5) {fox};
		\node [style=none] (42) at (2.75, 4.25) {};
		\node [style=none] (43) at (7.25, 4.25) {};
		\node [style=none] (44) at (5, 5.25) {};
		\node [style=none] (46) at (5, 4.5) {after};
		\node [style=none] (47) at (7.5, 4.25) {};
		\node [style=none] (48) at (9.5, 4.25) {};
		\node [style=none] (49) at (8.5, 5.25) {};
		\node [style=none] (51) at (8.5, 4.5) {chicken};
	\end{pgfonlayer}
	\begin{pgfonlayer}{edgelayer}
		\draw [in=90, out=-90] (1.center) to (2.center);
		\draw [in=90, out=-90] (4.center) to (5.center);
		\draw [in=90, out=-90] (7.center) to (8.center);
		\draw [in=90, out=-90] (10.center) to (11.center);
		\draw [in=90, out=-90] (13.center) to (14.center);
		\draw [in=90, out=-90] (16.center) to (17.center);
		\draw [in=90, out=-90] (19.center) to (20.center);
		\draw [in=90, out=-90] (22.center) to (23.center);
		\draw [in=90, out=-90] (25.center) to (26.center);
		\draw [in=180, out=-90] (5.center) to (28.center);
		\draw [in=0, out=-90] (11.center) to (28.center);
		\draw [in=180, out=-90] (2.center) to (29.center);
		\draw [in=0, out=-90] (14.center) to (29.center);
		\draw [in=180, out=-90] (8.center) to (30.center);
		\draw [in=0, out=-90] (17.center) to (30.center);
		\draw [in=180, out=-90] (23.center) to (31.center);
		\draw [in=0, out=-90] (26.center) to (31.center);
		\draw [-, fill=white] (32.center) to (33.center);
		\draw [-, fill=white] (33.center) to (34.center);
		\draw [-, fill=white] (37.center) to (38.center);
		\draw [-, fill=white] (38.center) to (39.center);
		\draw [-, fill=white] (42.center) to (43.center);
		\draw [-, fill=white] (43.center) to (44.center);
		\draw [-, fill=white] (47.center) to (48.center);
		\draw [-, fill=white] (48.center) to (49.center);
		\draw (42.center) to (44.center);
		\draw (32.center) to (34.center);
		\draw (37.center) to (39.center);
		\draw (47.center) to (49.center);
	\end{pgfonlayer}
\end{tikzpicture}
\\
{\ttfamily \scriptsize
\textbf{Target:} chases\\
\textbf{Prediction:} chases (1.00)}
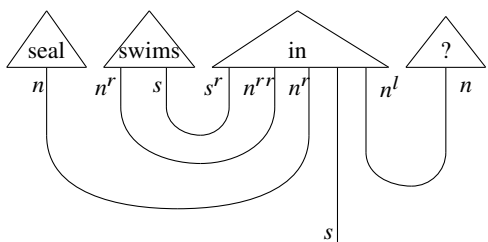
\vspace{10pt}

\begin{tikzpicture}[baseline={(0.base)}, scale=.3]
	\begin{pgfonlayer}{nodelayer}
		\node [style=none] (0) at (3, 2.25) {};
		\node [style=none] (1) at (9.5, 4.5) {};
		\node [style=none] (2) at (9.5, 0.75) {};
		\node [style=none, right] (3) at (10.1, 3.65) {$n$};
		\node [style=none] (4) at (-8, 4.5) {};
		\node [style=none] (5) at (-8, 1.5) {};
		\node [style=none, right] (6) at (-8.65, 3.65) {$n$};
		\node [style=none] (7) at (-4.75, 4.5) {};
		\node [style=none] (8) at (-4.75, 2.75) {};
		\node [style=none, right] (9) at (-5.9, 3.65) {$n^r$};
		\node [style=none] (10) at (-2.75, 4.5) {};
		\node [style=none] (11) at (-2.75, 2.75) {};
		\node [style=none, right] (12) at (-3.4, 3.65) {$s$};
		\node [style=none] (13) at (0, 4.5) {};
		\node [style=none] (14) at (0, 2.75) {};
		\node [style=none, right] (15) at (-1.15, 3.65) {$s^r$};
		\node [style=none] (16) at (2, 4.5) {};
		\node [style=none] (17) at (2, 2.75) {};
		\node [style=none, right] (18) at (0.6, 3.65) {$n^{rr}$};
		\node [style=none] (19) at (3.5, 4.5) {};
		\node [style=none] (20) at (3.5, 1.5) {};
		\node [style=none, right] (21) at (2.6, 3.65) {$n^r$};
		\node [style=none] (22) at (4.75, 4.5) {};
		\node [style=none] (23) at (4.75, -3.5) {};
		\node [style=none, right] (24) at (4.1, -2.85) {$s$};
		\node [style=none] (25) at (6, 4.5) {};
		\node [style=none] (26) at (6, 0.75) {};
		\node [style=none, right] (27) at (6.6, 3.65) {$n^l$};
		\node [style=none] (28) at (-1.25, 1.5) {};
		\node [style=none] (29) at (-1.25, 0.25) {};
		\node [style=none] (30) at (-2.25, -1.75) {};
		\node [style=none] (31) at (8, -0.75) {};
		\node [style=none] (32) at (7.75, 4.5) {};
		\node [style=none] (33) at (11.25, 4.5) {};
		\node [style=none] (34) at (9.5, 6.75) {};
		\node [style=none] (36) at (9.5, 5.25) {?};
		\node [style=none] (37) at (-9.75, 4.5) {};
		\node [style=none] (38) at (-6.25, 4.5) {};
		\node [style=none] (39) at (-8, 7) {};
		\node [style=none] (41) at (-8, 5.25) {seal};
		\node [style=none] (42) at (-5.5, 4.5) {};
		\node [style=none] (43) at (-1.5, 4.5) {};
		\node [style=none] (44) at (-3.5, 7) {};
		\node [style=none] (46) at (-3.5, 5.25) {swims};
		\node [style=none] (47) at (-0.75, 4.5) {};
		\node [style=none] (48) at (7, 4.5) {};
		\node [style=none] (49) at (3, 7) {};
		\node [style=none] (51) at (3, 5.25) {in};
	\end{pgfonlayer}
	\begin{pgfonlayer}{edgelayer}
		\draw [in=90, out=-90] (1.center) to (2.center);
		\draw [in=90, out=-90] (4.center) to (5.center);
		\draw [in=90, out=-90] (7.center) to (8.center);
		\draw [in=90, out=-90] (10.center) to (11.center);
		\draw [in=90, out=-90] (13.center) to (14.center);
		\draw [in=90, out=-90] (16.center) to (17.center);
		\draw [in=90, out=-90] (19.center) to (20.center);
		\draw [in=90, out=-90] (22.center) to (23.center);
		\draw [in=90, out=-90] (25.center) to (26.center);
		\draw [in=180, out=-90] (11.center) to (28.center);
		\draw [in=0, out=-90] (14.center) to (28.center);
		\draw [in=180, out=-90] (8.center) to (29.center);
		\draw [in=0, out=-90] (17.center) to (29.center);
		\draw [in=180, out=-90] (5.center) to (30.center);
		\draw [in=0, out=-90] (20.center) to (30.center);
		\draw [in=180, out=-90] (26.center) to (31.center);
		\draw [in=0, out=-90] (2.center) to (31.center);
		\draw [-, fill=white] (32.center) to (33.center);
		\draw [-, fill=white] (33.center) to (34.center);
		\draw [-, fill=white] (37.center) to (38.center);
		\draw [-, fill=white] (38.center) to (39.center);
		\draw [-, fill=white] (42.center) to (43.center);
		\draw [-, fill=white] (43.center) to (44.center);
		\draw [-, fill=white] (47.center) to (48.center);
		\draw [-, fill=white] (48.center) to (49.center);
		\draw (42.center) to (44.center);
		\draw (37.center) to (39.center);
		\draw (49.center) to (47.center);
		\draw (34.center) to (32.center);
	\end{pgfonlayer}
\end{tikzpicture}
\\
{\ttfamily \scriptsize
\textbf{Target:} water\\
\textbf{Prediction:} water (0.97), dog (0.02)}
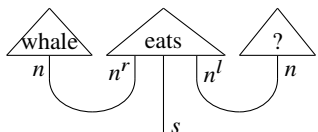
\vspace{10pt}

\begin{tikzpicture}[baseline={(0.base)}, scale=.25]
	\begin{pgfonlayer}{nodelayer}
		\node [style=none] (0) at (1, 2.25) {};
		\node [style=none] (1) at (7, 3.5) {};
		\node [style=none] (2) at (7, 2.25) {};
		\node [style=none, right] (3) at (7.35, 2.65) {$n$};
		\node [style=none] (4) at (-5, 3.5) {};
		\node [style=none] (5) at (-5, 2.25) {};
		\node [style=none, right] (6) at (-5.9, 2.65) {$n$};
		\node [style=none] (7) at (-0.5, 3.5) {};
		\node [style=none] (8) at (-0.5, 2.25) {};
		\node [style=none, right] (9) at (-1.9, 2.65) {$n^r$};
		\node [style=none] (10) at (1, 3.5) {};
		\node [style=none] (11) at (1, -0.75) {};
		\node [style=none, right] (12) at (1.35, -0.35) {$s$};
		\node [style=none] (13) at (2.75, 3.5) {};
		\node [style=none] (14) at (2.75, 2.25) {};
		\node [style=none, right] (15) at (3.1, 2.65) {$n^l$};
		\node [style=none] (16) at (-2.75, 0.5) {};
		\node [style=none] (17) at (5, 0.5) {};
		\node [style=none] (18) at (5, 3.5) {};
		\node [style=none] (19) at (9, 3.5) {};
		\node [style=none] (20) at (7, 6) {};
		\node [style=none] (22) at (7, 4.25) {?};
		\node [style=none] (23) at (-7.25, 3.5) {};
		\node [style=none] (24) at (-2.75, 3.5) {};
		\node [style=none] (25) at (-5, 6) {};
		\node [style=none] (27) at (-5, 4.25) {whale};
		\node [style=none] (28) at (-2, 3.5) {};
		\node [style=none] (29) at (4.25, 3.5) {};
		\node [style=none] (30) at (1, 6) {};
		\node [style=none] (32) at (1, 4.25) {eats};
	\end{pgfonlayer}
	\begin{pgfonlayer}{edgelayer}
		\draw [in=90, out=-90] (1.center) to (2.center);
		\draw [in=90, out=-90] (4.center) to (5.center);
		\draw [in=90, out=-90] (7.center) to (8.center);
		\draw [in=90, out=-90] (10.center) to (11.center);
		\draw [in=90, out=-90] (13.center) to (14.center);
		\draw [in=180, out=-90] (5.center) to (16.center);
		\draw [in=0, out=-90] (8.center) to (16.center);
		\draw [in=180, out=-90] (14.center) to (17.center);
		\draw [in=0, out=-90] (2.center) to (17.center);
		\draw [-, fill=white] (18.center) to (19.center);
		\draw [-, fill=white] (19.center) to (20.center);
		\draw [-, fill=white] (23.center) to (24.center);
		\draw [-, fill=white] (24.center) to (25.center);
		\draw [-, fill=white] (28.center) to (29.center);
		\draw [-, fill=white] (29.center) to (30.center);
		\draw (25.center) to (23.center);
		\draw (28.center) to (30.center);
		\draw (20.center) to (18.center);
	\end{pgfonlayer}
\end{tikzpicture}
\\
{\ttfamily \scriptsize
\textbf{Target:} krill\\
\textbf{Prediction:} cheese (0.53), fish (0.17), grain (0.10)}

\vspace{12pt}

\begin{figure}[h]
{\ttfamily \tiny \input{appendix/test_sentences.tex}}
\caption{Testing set (23 sentences)}
\end{figure}

\begin{figure}[h]
{\ttfamily \tiny \input{appendix/train_sentences.tex}}
\caption{Training set (86 sentences)}
\end{figure}

\end{document}